\documentclass{article}
\usepackage{spconf,amsmath,epsfig}
\usepackage{amssymb}
\usepackage{multirow}
\usepackage{makecell}
\usepackage{relsize}
\usepackage{xcolor}
\usepackage{amsmath}
\usepackage{enumitem}

\let\OLDthebibliography\thebibliography
\renewcommand\thebibliography[1]{
  \OLDthebibliography{#1}
  \setlength{\parskip}{0pt}
  \setlength{\itemsep}{0pt plus 0.3ex}
}

\begin{document}\sloppy


\title{\textit{AM$^2$-EmoJE}:\textit{A}daptive \textit{M}issing-\textit{M}odality \textit{Emo}tion Recognition in Conversation via \textit{J}oint \textit{E}mbedding Learning}
%
\name{Naresh Kumar Devulapally, Sidharth Anand, Sreyasee Das Bhattacharjee, Junsong Yuan}
\vspace{-2cm}
\address{The State University of New York at Buffalo}

\maketitle

\begin{abstract}
Human emotion can be presented in different modes i.e., audio, video, and text. However, the contribution of each mode in exhibiting each emotion is not uniform. Furthermore, the availability of complete mode-specific details may not always be guaranteed in the test time. In this work, we propose \textit{AM$^2$-EmoJE}, a model for \textit{A}daptive \textit{M}issing-\textit{M}odality \textit{Emo}tion Recognition in Conversation via \textit{J}oint \textit{E}mbedding Learning model that is grounded on two-fold contributions: \textbf{First}, a \textit{query adaptive fusion} that can automatically learn the relative importance of its mode-specific representations in a query-specific manner. By this the model aims to prioritize the mode-invariant spatial (within utterance) query details of the emotion patterns, while also retaining its mode-exclusive aspects within the learned multimodal query descriptor. 
\textbf{Second} the \textit{multimodal joint embedding learning} module that explicitly addresses various missing modality scenarios in test-time. By this, the model learns to emphasize on the correlated patterns across modalities, which may help align the cross-attended mode-specific descriptors pairwise within a joint-embedding space and thereby compensate for missing modalities during inference. By leveraging the spatio-temporal details at the dialogue level, the proposed  \textit{AM$^2$-EmoJE} not only demonstrates superior performance (around $2-4\%$ improvement in Weighted-F1 scores) compared to the best-performing state-of-the-art multimodal methods, by effectively leveraging body language in place of face expression, it also exhibits an enhanced privacy feature. By reporting around $2-5\%$ improvement in the weighted-F1 score, the proposed multimodal joint embedding module facilitates an impressive performance gain in a variety of missing-modality query scenarios during test time.

\end{abstract}
%
\section{Introduction}
\label{sec:intro}

\begin{figure}[ht]

\centerline{\includegraphics[width=0.45\textwidth]{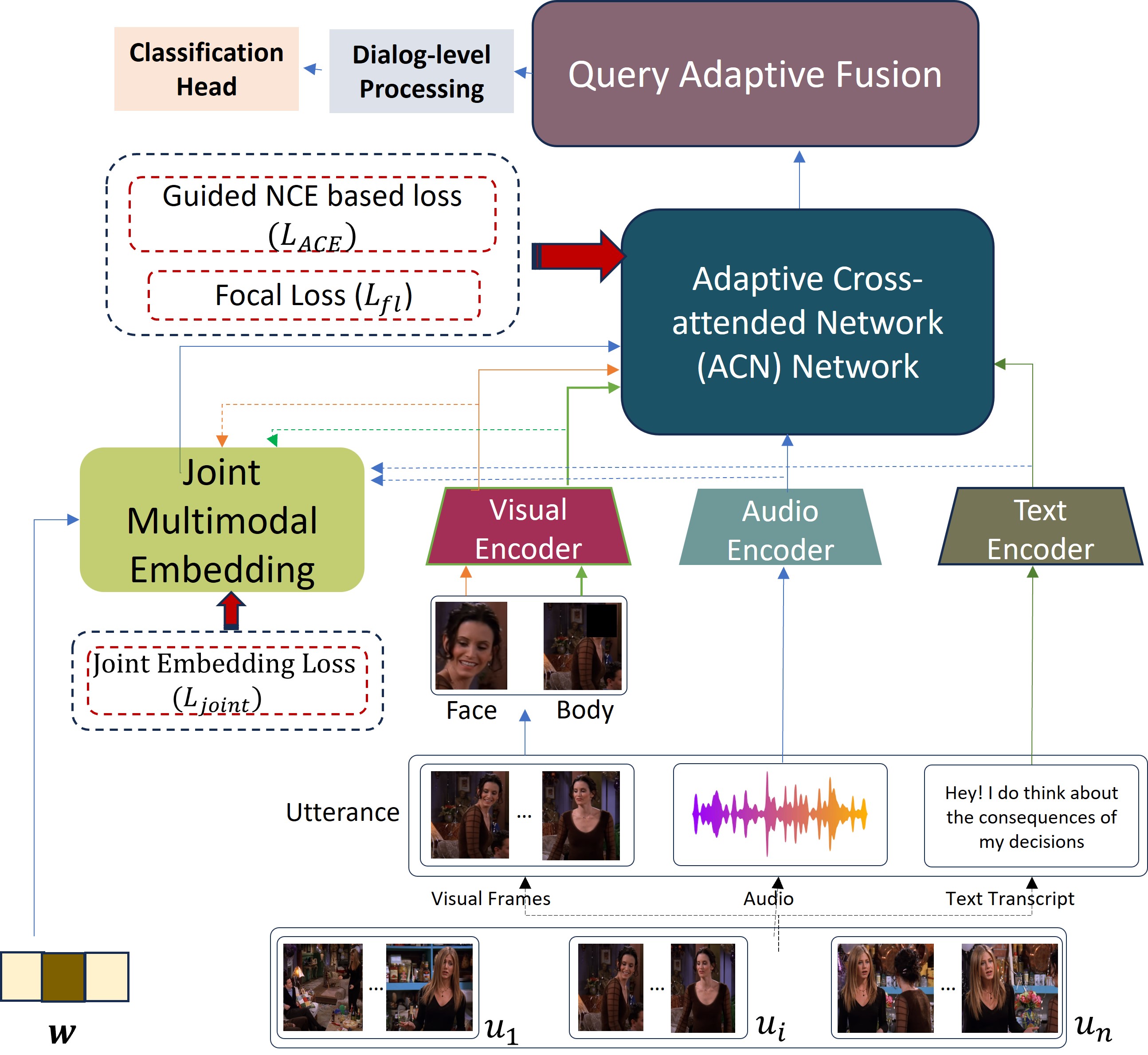}}
\caption{\small{Proposed \textit{AM$^2$-EmoJE} framework}}
\centering
\label{fig:arch}
\vspace{-8mm}
\end{figure}
\vspace{-2mm}
Emotion analysis and its dynamic evolution in humans during conversations is pivotal to various critical tasks ranging from sentiment analysis in social media to affect-aware human-robot interactions. The problem is particularly complex due to the involvement of multi-party stakeholders, their mutual interactions described via different modalities like text, video, and audio, as well as their body language, which influence the speaker's emotions and its spatio-temporal progression expressed in an utterance. 

In particular, a speaker's emotion depends on various intra (e.g. personal details, behavioral patterns, and habits) and inter (e.g. audience behavior, their interpreting conducts) personal contexts and other environmental circumstances. At one end, most multimodal methods assume data completeness, which may
not hold in practice due to privacy, device, or security constraints. We note that the mode-specific descriptor components exhibiting an individual's emotion have some correlation, which may be instrumental to handling the missing modality information. However, this aspect has not yet been sufficiently explored by the existing literature.
At the other end, the significance of this variety of heterogeneous mode information may not be uniform in evaluating emotions for all speakers. For example, a speaker's facial expression may have only limited information about their true emotion, while their body language may still be more semantically rich. 
Thus the challenges related to \textit{the presence of strong heterogeneity both within and across multiple modalities and complex interplay} and \textit{the influence of a missing modality information toward learning comprehensive multimodal interaction patterns in a real-life conversation setting} are critical yet under-studied.

Toward these, we propose \textit{AM$^2$-EmoJE}, an \textit{A}daptive \textit{M}issing-\textit{M}odality \textit{Emo}tion Recognition in Conversation via \textit{J}oint \textit{E}mbedding Learning model that is grounded on the following contributions. 

\vspace{-1mm}
\begin{enumerate}[leftmargin=0.2cm,itemindent=0cm,labelwidth=\itemindent,align=left]

   \vspace{-2mm}
    \item \textit{Query Adaptive Fusion (QAF)} that derives the relative importance of the cross-attended mode descriptors on the fly to deliver a robust multimodal query descriptor preserving both 
   instance-specific and category-specific utterance-level spatial patterns in parallel.

   \vspace{-2mm}
   \item {\textit{Multimodal joint embedding}} that may leverage an auxiliary boolean mask vector as user input to turn on an effective mode switching mechanism, which proves instrumental in delivering a competitive decision ability for queries with incomplete mode information.

   \vspace{-2mm}
    \item \textit{Extensive Evaluation Analysis} using publicly available (MELD \cite{poria2019meld}, IEMOCAP \cite{Busso2008}) datasets not only demonstrate an impressive classification performance ($2-4\%$ improvement in Weighted-F1 of \textit{AM$^2$-EmoJE} in complete multimodal test scenarios, but also exhibit a competitive decision making for queries with missing modality information.

\end{enumerate}

\section{Related Work}


Multimodal solutions toward motion Recognition in Conversations (ERC)  have recently demonstrated significant performance improvements compared to their unimodal counterparts. While natural language transcriptions serve as strong emotion indicators for unimodal ER pipelines, leveraging the advantages of cross-modal interactions demonstrates significant performance gain\cite{chudasama2022m2fnet, li-etal-2022-emocaps,kiela2018efficient,shi2022learning}.

While exhibiting promising performances, traditional multimodal methods assume data completeness, which may
not be a feasible constraint for various practice issues related to privacy, device, or security constraints and missing modality in a test environment appears to be challenging to these models. In a recent work, Ma et al. \cite{9879085} investigate the behavior of Transformer-based multimodal frameworks in the presence of modal-incomplete data. Lee et al. \cite{Lee2023MultimodalPW} propose a prompt learning framework to address the issues of missing modalities during training or testing. However, the limitation of computational resources still persists as a bottleneck to such training-heavy transformer models. 

While a few recent works \cite{LIU2024101973,10.1145/3477495.3532064,10.1609/aaai.v37i2.25253, 10.1145/3534678.3539388} attempt to address the missing modality challenge in specific application settings, a straightforward extension to these preliminary models for the complex task of ERC, may not be feasible. Evaluating the relative contribution of multiple mode-specific components describing a speaker's emotion dynamics on the fly, while estimating their missing-mode description in parallel, is challenging. Toward these, we propose \textit{AM$^2$-EmoJE} that is motivated by two-fold research objectives:\textbf{First}, a \textit{query adaptive fusion} that can automatically learn the relative importance of its mode-specific representations in a query-specific manner; \textbf{Second} \textit{multimodal joint embedding learning} framework that learns a mode-switching mechanism to align the cross-attended mode-specific descriptors pairwise within a joint-embedding space and thereby allowing the model to compensate for missing modalities during inference.

\vspace{-4mm}
\section{Proposed Method}
\textbf{Problem Definition}: Given a multi-party dialogue $d:=\{u_j\}_j\in \mathcal D$ represented as a sequence of utterances $\{u_j\}_j$, the objective is to evaluate the dominant emotional states of the speaker as expressed in each utterance $u_j$. For brevity, now onward we will omit the suffix $j$, and an arbitrary utterance $u_j$ will be represented as $u$ unless the suffix is specifically required. Each $u \in \mathcal{D}$ contains $(u_v, u_a, u_t)$, where $u_v$ is the \textit{video}, $u_a$ is the \textit{audio}, and  $u_t$ is the \textit{text transcription} component of the utterance. In this work, we propose a continual emotion evaluation model that can evaluate the series of dialogues by single or multiple speakers in a conversational environment. Unlike existing multimodal literature, which primarily relies on the availability of complete mode-specific components representing a multimode utterance, in this work we allow incomplete query videos, i.e., some of the mode-specific information related to the query may be missing. Therefore, the availability information of each mode-specific query component is passed by a \textit{boolean mask vector} $\mathbf w:=(w_v, w_a, w_t)$. For example, a query that only has an audio and transcript will is passed with a mask vector $\mathbf w$ with $w_v=0$. In another query environment, if a user is only interested in analyzing the visual component of a multimode query, then the mask vector $\mathbf w$ with $w_a=0$ and $w_t=0$ is passed with the query that may enable the system to perform only the visual analysis.


\vspace{-4mm}
\subsection{Mode-Specific Feature Representation}

\vspace{-2mm}
To capture the spatio-temporal evolution of information within each utterance, the first-level mode-specific feature representation scheme as described below:


\vspace{-6mm}
\subsubsection{Text Representation}

\vspace{-2mm}
To derive a compact descriptor for the text component $u_t$ represented as a sequence of $w$ words, i.e. $u_t=\{\omega_1, \omega_2, ..., \omega_w\}$, we employ the pretrained model SBERT \cite{sbert} to obtain the fixed language embedding  $\mathbf f^{T}\in \mathbb R^{w\times d_{t}}$ for the text component $u_t$.


\vspace{-5mm}
\subsubsection{Video Representation}

\vspace{-2mm}
 For the visual component $u_v$ of each utterance $u\in d$, FFmpeg is used to identify $n$ key frames and MTCNN \cite{zhang2016joint} is applied to extract the aligned faces from each key frame. To deliver a comprehensive analysis of the subjects' emotional evolution to appear in an utterance, we analyze their facial expression in contrast to their body language, each $u_v$ is split into two sequences, which cover the visuals of distinct regions in the utterance: `face sequence' that contains the sequence of derived frames containing only the subjects' face regions; `body sequence' that exclusively contains the subjects' body segment in the utterance. YOLOv7 \cite{wang2023yolov7} fine-tuned for human detection, is employed to localize the human body segment in a frame, To analyze the body language in exclusion, we subtract the face region from the identified body segment in a frame to design the derived frames in the `body sequence'. 
 Thus, the visual content $u_v$ of $u$ is represented in terms of two equal-sized derived frame sequences: $\mathbf v^{face}=\{\mathbf{au}_{1}, \mathbf{au}_{2}, ..., \mathbf{au}_{n}\}$ and $\mathbf v^{body}=\{\mathbf b_1, \mathbf b_2, ..., \mathbf b_n\}$, where each $\mathbf{au}_{j}$ and $\mathbf b_j$ represent a learned descriptor describing the $j^{th}$ element in $\mathbf v^{face}$ and $\mathbf v^{body}$ respectively. Two identical Bi-LSTM-based sequence representation modules with a hidden embedding size of $d_v$, which take $\mathbf v^{face}$ or $\mathbf v^{body}$ as inputs, are employed to obtain the initial regional descriptors $\mathbf f^{face}\in \mathbb R^{n\times d_v}$ or $\mathbf f^{body}\in \mathbb R^{n\times d_v}$. These two independent LSTM networks are then merged via a stacked self-attention layer that attends to the pair of these network-specific inputs to derive a self-attended visual descriptor  $\mathbf f^{v}\in \mathbb R^{2n\times d_{v}}$ for $u_v$.


\vspace{-4mm}
\subsubsection{Audio Representation}

\vspace{-2mm}
We use the Pathout Fast 2D Spectrogram model (PASST) \cite{koutini22passt}, which is initialized from a vision transformer trained on ImageNet, to fine-tuned on 10s audio clips from AudioSet. To derive a fine-grained understanding of the audio signal evolution, the audio component of each utterance is divided into smaller $e$ overlapped segments and each segment is represented using their corresponding PASST descriptor $\mathbf a_i\in \mathbb R^{d_{passt}}$ such that $u_a:==\{\mathbf a_1, \mathbf a_2, ..., \mathbf a_e\}$. Similar to the video representation, we use a BiLSTM to derive an overall audio feature for the utterance $\mathbf f^a\in\mathbb{R}^{d_a}$.

\vspace{-4mm}
\subsection{Weighted Multimodal Attention}

\vspace{-2mm}
Toward integrating the mode-specific contents across modalities, we note that the expression captured by these uni-mode components may not solely reflect their intra-mode data pattern. They may also represent the influence of information depicted by other modalities. For example, a speaker's audio may reflect certain emotions, which may not be sufficiently supported by their body language or an utterance transcript may be continually modulated by the expressions shown by the audience in the previous utterance segments. Furthermore, the availability of a quality mode-specific representation of an utterance may also not be guaranteed. For instance, the availability of visual cues may not be assured in every query scenario. Otherwise, noisy audio may also deteriorate the overall data quality occasionally. 

While existing multimodal emotion analysis models do not allow such incomplete query information, we design an Adaptive Cross-attended Network (ACN) that enables a flexible yet robust feature representation technique to capture the effects of same- and different-modality cues at various levels of details (including missing modality cues).
As observed in Figure \ref{fig:arch}, each layer of \textit{ACN} is administered by multiple cross-modal cues from multiple \textit{outer} networks into a single mode-specific \textit{inner} network. The merging layer of \textit{ACN} incorporates a simple yet effective masking mechanism toward facilitating the learning of several variants of mode combinations, wherein the model is explicitly trained to control the contribution of each 'outer' network including a variety of missing-modality scenarios. The keys and values are generated as $\mathbf K^{l}_{m_i}=linear(w_{m_i}\mathbf (W^{l, K}_{m_i})^{T}\mathbf {e}_{m_i})$, $\mathbf V^{l}_{m_i}=linear(w_{m_i}\mathbf (W^{l, V}_{m_i})^{T}\mathbf {e}_{m_i})$, where $W^{l,K}\in\mathbb{R}^{d_{m_i}\times d_l}$ and $W^{l,V}\in\mathbb{R}^{d_{m_i}\times d_l}$ are key and value weight matrices in the $l^{th}$ layer of the attention network. The output of each attention head in the $l^{th}$ layer is then computed as:
\begin{equation}
\scriptsize
    \mathbf g^l_{m}=\mathbf g^{l-1}_{m}+\frac{1}{|\mathcal M|}\mathrm{softmax}(linear(\frac{\mathbf g^{l-1}_{m}(\sum_{m_{i}\in \mathcal M\setminus \{m\} }\mathbf K^{l-1}_{m_i})^T}{\sqrt{d_{m_i}}}))\mathbf V^{l-1}_{m_i}
\end{equation}

Thus, for each modality $m$ in $\mathcal{M}$, the set of all modalities leveraged in the model, the learned descriptor $\mathbf g^m$ from multiple heads are mean-pooled to capture the weighted aggregated details within the final output of the \textit{Central} network.


As shown in Figure 1, each mode-specific \textit{Central} query network for each $m\in \mathcal{M}$ of  \textit{ACN} thus produces an average pooled cross-attended mode-specific descriptor $\mathbf f^{m}_{ACN}\in \mathbb R^d$ for the uni-mode components $t$, $v$, and $a$ for $u$.  We will discuss the learning algorithm later in Section \ref{learn_obj}.

\vspace{-4mm}
\subsection{Query Adaptive Fusion}

\vspace{-2mm}
In contrast to the existing fusion techniques, which assume that the mode-specific contributions in the resulting multimodal descriptor for a query input should be uniform, in reality, the quality and availability of the complete mode-specific components may not be feasible in general. Therefore, we propose a \textit{Q}uery \textit{A}daptive \textit{F}usion (\textit{QAF}) mechanism that designs a linear combination of the learned cross-attended mode descriptors $\mathbf g^{m}$ to define a comprehensive multimode utterance-level descriptor as follows:

\vspace{-5mm}
\begin{equation}
    \mathcal A(u) = \sum_{m \in \mathcal{M}}\left(\frac{w_m}{|\mathcal{M}|}\sum_{m_i\in\mathcal M\setminus m}\alpha^m_{m_i}\mathbf f^{m_i}+ (1 - \alpha^m_{m_i})\mathbf f^{m_i}\right)
\end{equation}
where $0 \leq\alpha^m_{m_i} \forall m,m_i \in \mathcal M\leq 1$ are learnable parameters. 

Thus, the proposed fusion function $\mathcal A$ provides a flexible multimodal representation mechanism, by which the resulting multimodal descriptor $\mathcal A(u)$ for an utterance $u$ can retain category-specific discriminative data patterns, however not completely disregarding the unique instance-specific data patterns observed in the utterance. Furthermore, given an utterance $u$, in the absence of its complete mode-specific details (as provided by the user-specified mask vector $\mathbf w$) related to $m$, the model automatically learns to adopt its learnable parameters $\{\alpha^m_{m_i}\}_{m_i}$ to derive an adjusted multimodal descriptor $\mathcal A(u)$. We adopt the optimization approach of \cite{alfamix} to learn the interpolation parameters $\alpha^m_{m_i}$.

\vspace{-5mm}
\subsection{Classification}

\vspace{-2mm}
Given a conversational dialogue represented using a sequence of $n$ utterances $\{u_j\}^{n}_j\in \mathcal D$, the emotion of the speaker $s$ is estimated by leveraging two parallel utterance sequences: \textit{Dialogue Context} that describes the sequence $\{\mathcal A(u_j)\}_j$; \textit{Speaker Context} that describes a sub-sequence $\{\mathcal A(u_{{s}_j})\}_j$, where the sub-sequence $\{u_{{s}_j}\}_{{s_{j}\in [1, n]}}$ is generated from the dialogue and includes only those utterances, in which $s$ vocally contributes to the conversation. Two parallel Bi-LSTMs are trained to capture the spatio-temporal contexts independently of these contexts' perspectives: $\mathbf s_l \in \mathbb R^{s}$ representing the \textit{Speaker Context} and  $\mathbf d_l \in \mathbb R^{s}$ representing the \textit{Dialogue Context}. These representations are passed through a classification head to arrive at the classification decision for the emotion for the utterance $u$.

\vspace{-4mm}
\subsection{Training objectives}
\label{learn_obj}
\vspace{-2mm}
The learning of \textit{AM$^2$-EmoJE} includes two independent
learning objectives: a multi-component loss objective ($\mathcal L$) that combinedly optimizes the guided NCE loss ($\mathcal{L}_{ACE}$) and the focal loss ($\mathcal{L}_{fl}$) for the classifier; a CLIP-like loss function $\mathcal L_{joint}$ to learn the pairwise joint embedding spaces. We discuss them below.

\vspace{-4mm}
\subsubsection{Guided NCE}

\vspace{-2mm}
The proposed Adaptive Cross-attended Network (\textit{ACN}) jointly learns the cross-attended representations $\mathbf f^{t}_{ACN,j}$, $\mathbf f^{v}_{ACN,j}$, and $\mathbf f^{a}_{ACN,j}$ with twofold contributions: 1) preserving instance specific discriminability by identifying more reliable modes on the fly. We intuitively expect a speaker's face and body language to display similar emotions, and a significant deviation from this would require us to rely more on the other modalities like audio or transcript to accurately estimate their true emotions. Guided by this observation, we leverage an aggregated noise contrastive estimation ($\mathcal L_{ACE}$) as below:

\vspace{-6mm}
\begin{equation}
\label{fl_loss}
    \mathcal{L}_{ACE}=\frac{1}{|\mathcal D|}\sum_{u_{j}\in \mathcal D}\frac{1}{|\mathcal M|}\sum_{\substack {m\neq m_{i}\\ m, m_{i}\in \mathcal M }}{\mathcal{L}_{\mathrm{NCE}}(\mathbf f^{m}_{ACN,j}, \mathbf f^{m_i}_{ACN,j})} 
    \vspace{-2mm}
\end{equation}
with

\vspace{-4mm}
\footnotesize
\begin{multline}
\mathcal{L}_{\mathrm{NCE}}(\mathbf{f}^{m}_{ACN,j}, \mathbf{f}^{m_{i}}_{ACN,j}) = \\
\Bigg[-\log\Bigg(\frac{P(\mathbf{f}^{m}_{ACN,j}|\mathbf{f}^{m_i}_{ACN,j})}{P(\mathbf{f}^{m}_{ACN,j}|\mathbf{f}^{m_i}_{ACN,j}) + \frac{|\mathcal{N}_j|}{|\mathcal{N}|}}\Bigg) \\
+ \sum_{k\in \mathcal{N}_{j}}\log\Bigg(\frac{P(\mathbf{f}^{m}_{ACN,k}|\mathbf{f}^{m_i}_{ACN,j})}{P(\mathbf{f}^{body}_{ACN,k}|\mathbf{f}^{face}_{ACN,j})+\frac{|\mathcal{N}_j|}{|\mathcal{N}|}}\Bigg) - 1\Bigg]
\end{multline}

that computes the 
probability of both features $\mathbf f^{m}_{ACN,j}$ and $\mathbf f^{m_i}_{ACN,j}$ representing the same instance $u_j$ compared to other elements in a uniformly sampled negative set $\mathcal N_j$. $\mathcal N$ is the sample batch.
The averaged Focal Loss $\mathcal L_{fl}$ \cite{focal}, specifically effective for an imbalanced dataset like ours, is employed to preserve the category details and a combined loss function $\mathcal L=\mathcal L_{ACE}+\mathcal L_{fl}$ to jointly learn its mode-specific \textit{Central} query networks.


\vspace{-4mm}
\subsubsection{Multimodal Joint Embedding} 

\vspace{-2mm}
To address the potential missing modality scenarios in a test environment, we introduce an effective joint latent representation learning model that follows a similar approach as proposed by Radford  et al. \cite{radford2021learning}: the latent descriptor for a positive pair (i.e. a pair of mode-specific representatives describing the same sample using two different modes) are mapped closely, while a negative pair (i.e. a pair of mode-specific representatives describing the two different samples using two different modes) samples will be mapped father from each other in the joint embedding space. Formally, given a minibatch of $\mathcal N$ of samples represented using their cross-attended descriptors in a two-mode space (i.e.  $\mathcal N:=\{(\mathbf f^{m_k}_i,\mathbf f^{m_l}_i )\}_{i, m_k\neq m_l}$), we use Jensen-Shannon Divergence, which is a symmetric version of the KL-Divergence to learn the pair-wise invertible linear mappings $S_{m_k\rightarrow m_{joint}}:\mathcal R^{d_{m_k}}\rightarrow \mathcal R^{d_{joint}}$, $\forall m_k\in \mathcal M$  to a single joint embedding space $\mathcal D_{joint}\subset \mathcal R^{d_{joint}}$ and the corresponding loss function is defined as:

\vspace{-6mm}
\begin{multline}
    \mathcal L_{m_k\rightarrow m_{joint}}=\frac{-1}{|\mathcal N|}\big[\sum_{x\in\mathcal N} \frac{1}{2}\log(\mathrm{KL}(S_{m_k\rightarrow m_{joint}}(\mathbf f^{m_k}_i)||\mathbf t^{kl}_i)\\+ \log(\mathrm{KL}(S_{m_l\rightarrow m_{joint}}(\mathbf f^{m_l}_i)||\mathbf t^{kl}_i)\big]
\end{multline}

where $x:=(\mathbf f^{m_k}_i,\mathbf f^{m_l}_i )$ and $\mathbf t^{kl}_i=\frac{S_{m_k\rightarrow m_{joint}}(\mathbf f^{m_k}_i+\mathbf f^{m_l}_i)}{2}$. The model jointly optimizes a single loss function $\mathcal L_{joint}:=\underset{\substack {m_k, m_l\in \mathcal M \\ m_k\neq m_l}}{\sum}  \mathcal L_{m_k\rightarrow m_{joint}}$ to learn the collection of invertible mapping functions: $\{S_{m_k\rightarrow m_{joint}}\}_{m,k\in \mathcal M}$.

\section{Experiments}

\begin{table}[t]  
\footnotesize
\caption {\small{Performance Comparison of different methods using the weighted average F1 measure (W-Avg F1) on the MELD dataset with uni modal (T-Text, A-Audio, and V- Video) and multi-modal representation. Due to the imbalanced class distribution of the dataset, the `Fear' and `disgust' classes are represented as the minority classes, the proposed method was also compared against other $5$ majority classes (`Neutral', `Surprise', `Sadness', `Joy', and `Anger') in the dataset and the results are reported in column `w-avg F1 5 CLS'}. More details on emotion-specific comparison are provided in the supplementary material.}
\centering
\smaller 
\label{tab:meld}
\begin{tabular}{|p{3cm}|c|c|c|}
\hline
\multirow{2}{*}{Method} & \multirow{2}{*}{Mode} & \multirow{2}{*}{w-Avg F1} & w-avg F1 \\
 & & & 5-CLS \\ \hline
MFN \cite{zadeh2018memory} & T + A & 0.547 & 0.5732\\ \hline
\multirow{3}{*}{ICON \cite{hazarika-etal-2018-icon}} & T & 0.546 & 0.5718\\
 & A & 0.377 & 0.3947\\
 & T + A & 0.563 & 0.5897\\ \hline
\multirow{3}{*}{DialogueRNN \cite{majumder2019dialoguernn}} & T & 0.551 & 0.5759\\
 & A & 0.34 & 0.3542\\
 & T + A & 0.57 & 0.5971\\ \hline
\multirow{3}{*}{ConGCN \cite{ijcai2019p752}} & T & 0.574 & 0.5969\\
 & A & 0.422 & 0.44\\
 & T + A & 0.594 & 0.6175\\ \hline
DialogueCRN \cite{hu2021dialoguecrn} & T + A & 0.6073 & -\\ \hline
EmoCaps \cite{li2022emocaps} & T + A + V & 0.6400 & -\\ \hline
M2FNet \cite{chudasama2022m2fnet} & T + A + V & 0.6785 & -\\ \hline
Cross-Modal Distribution Matching \cite{liang2020semi} & T + A & 0.571 & -\\ \hline
Transformer Based Cross-modality Fusion \cite{xie2021robust} & T + A + V & 0.64 & -\\ \hline
Hierarchical Uncertainty \cite{chen2022modeling} & T + A + V & 0.59 & -\\ \hline
Shape of Emotion \cite{agarwal2021shapes} & T + A + V & 0.63 & -\\ \hline
UniMSE \cite{hu2022unimse} & T + A + V & 0.66 & -\\ \hline
EmotionCLIP \cite{zhang2023learning} & T + A + V & 0.3459 & -\\ \hline
\multirow{8}{*}{\textit{AM$^2$-EmoJE}} & T + A & 0.6263 & 0.6845 \\ 
 & T + V & 0.6196 & 0.6782\\ 
 & A + V & 0.5285 & 0.5825\\ 
 & T + A (JE) & 0.6836 & 0.7051\\ 
 & T + V (JE) & 0.6897 & 0.7106\\ 
 & A + V (JE) & 0.6085 & 0.6572\\ 
 & No Face & 0.6914 & 0.7944\\ 
 & No Body & 0.7089 & 0.8117\\ 
 & T + A + V & \textbf{0.7198} & \textbf{0.8205} \\ \hline
\end{tabular}
\end{table}
\vspace{-4mm}
\subsection{Datasets}
MELD \cite{poria2019meld} is a multimodal group conversational dataset that sources clips from the TV Series FRIENDS and consists of the 6 basic emotions. IEMOCAP \cite{li2022emocaps} is a dual-party conversational dataset acted out by professional actors and features conversations that are far longer on average than those in MELD.

\vspace{-4mm}
\subsection{Results}

\vspace{-2mm}
Tables \ref{tab:meld} and \ref{tab:iemocap} report the comparative performance of the proposed method against several state-of-the-art methods 
\cite{hazarika-etal-2018-icon,zadeh2018memory,majumder2019dialoguernn,hu2021dialoguecrn,li2022emocaps,chudasama2022m2fnet,liang2020semi,Yang2021HybridCL, xie2021robust,chen2022modeling, agarwal2021shapes,hu2022unimse} in the MELD and IEMOCAP datasets, respectively. While text appears as the most reliable unimodal feature, combining multiple features has been undoubtedly helpful to\textbf{ deliver a competitive $71.98\%$ weighted 
F1-score (w-avg F1), which is an improvement of $\approx 4\%$ across all classes, as compared to the best-performing baseline M2FNet\cite{chudasama2022m2fnet}.} MELD is an unbalanced dataset that has `fear' and `disgust' identified as the minority classes, with less than $50$ samples as representatives. \textit{AM$^2$-EmoJE} reports $82.05\%$ w-avg F1, which is \textbf{a significant $20\%$ improvement in the w-avg F1 score compared to the best-performing baseline using the five majority classes.} Compared to the existing works, which offer equal emphasis on all modes, adaptively identifying the more reliable modes to design the relative weight assignments in a query-specific manner has been proven to be effective for delivering a more accurate estimate of a speaker's emotional state. Furthermore, from the performance reported in the sub-row described using  the mode `T+A+V(no face)' (that uses cues from audio, transcript, and visual capturing subjects' body regions only), we observe that the proposed \textit{AM$^2$-EmoJE} still remains competitive (i.e., reporting $\approx 2\%$ improved w-avg F1 score) against the best-performing baseline. In fact, as we compare the result reported in another sub-row described using mode `T+A+V(no body)' (that uses cues from audio, transcript, and visual capturing subjects' face regions only), we observe that \textit{AM$^2$-EmoJE}'s performance remains nearly equivalent as in two different multi-mode data scenarios. Finally, to evaluate the contribution of the proposed multimodal joint learning (Section 3.5.2), in the table we also report the performance of the proposed \textit{AM$^2$-EmoJE} in multiple missing modality scenarios. \textbf{For example, comparing the performance between the sub-row described using mode `T+A' (that uses only text and audio) and `T+A(JE)' (that uses only text and audio and uses multimodal joint learning to compensate for the visual modality) we note that the proposed multimodal joint learning module enables around $5\%$ (and $2\%$) gain in the w-avg F1 score overall (and five majority) classes.} A similar observation can also be made by comparing the subrows `A+V' and `A+V(JE)' (and subrows  `T+V' and `T+V(JE)' ) which reports around $8\%$ (and $7\%$) performance gain in the w-avg F1 score. Augmented with the proposed multimodal joint learning module, \textit{AM$^2$-EmoJE} attains a state-of-the-art performance in two-mode data scenarios, where the queries are presented with a variety of missing modality scenarios (e.g. `video' or `audio').

A similar performance was also observed in Table \ref{tab:iemocap} which reports the comparative analysis using the IEMOCAP dataset. The proposed \textit{AM$^2$-EmoJE} \textbf{reports $74.91\%$ w-avg F1, which is around $4.9\%$ improvement compared to the best-performing baseline LIGHT-SERNET\cite{9746679}}. Being equipped with the proposed multimodal joint learning module, the model again demonstrates a competitive performance in two-mode data scenarios, where the queries are represented using `text and audio' or `text and video'.

Table \ref{tab:ablation} reports an ablation study. 
The first two rows in the Table demonstrate the superiority of the weighted multimodal attention (Section 3.2) and weighted fusion techniques (Section 3.3) compared to vanilla cross-attention and static weights (where we use equal weight for all modes). The subrows in the third row report the performance of the proposed \textit{AM$^2$-EmoJE} where only classification focal loss $\mathcal L_{fl}$ is used, but we did not use $\mathcal L_{ACE}$ in Eqn (3). \textbf{As observed in the table, while the introduction of the Guided NCE (Section 3.5.1) with focal loss delivers around $1-2\%$ improvement in w-F1 score,  by enabling a query adaptive multimodal fusion scheme (Section 3.3) \textit{AM$^2$-EmoJE} delivers a robust ($2-4\%$ improved) the w-F1 score.}

\vspace{-6mm}
\begin{table}[t]
\caption {Performance comparison of difference methods using the weighted average F1 measure (W-Avg F1) on the IEMOCAP dataset with uni (T:=Text, A:=Audio, and V:= Video) and multi-modal Data Representations. `Feature Concat' in row 13 and row 14 describe the concatenation of multiple uni-mode descriptors to define a multimodal descriptor. More details on emotion-specific comparison are provided in the supplementary material}
\label{tab:iemocap}
\centering
\small{
\begin{tabular}{|c|c|c|}
    \hline
       \textbf{Method} & \textbf{Mode} & \textbf{w-Avg F1}\\     \hline
      \makecell{MFN\cite{zadeh2018memory}} & T + A  & 0.3490\\ \hline
      \makecell{ICON\cite{hazarika-etal-2018-icon}} & T + A + V &  0.6350\\ \hline 
      \makecell{DialogueRNN\cite{majumder2019dialoguernn}} & T + A + V & 0.6275\\ \hline
      \makecell{MMGCN\cite{10.1145/3343031.3351034}} & T + A + V &  0.6622\\ \hline
      \makecell{DialogueCRN\cite{hu2021dialoguecrn}} & T + A  & 0.6620\\ \hline
              \makecell{Hierarchical Uncertainty \\ \cite{chen2022modeling} }& T + A + V & 0.6598\\ \hline
      \makecell{DAG-ERC+HCL\cite{Yang2021HybridCL}} & T  & 0.6803\\ \hline
       \makecell{M2FNet\cite{chudasama2022m2fnet}} & T + A + V & 0.6986\\ \hline
        \makecell{LIGHT-SERNET\cite{9746679}} & T + A + V& 0.7020\\ \hline

        \multirow{8}*{\textit{AM$^2$-EmoJE}} & T + A  & 0.6162\\ \cline{2-3}
            & T + V & 0.6343\\ \cline{2-3}
            & A + V & 0.5379\\ \cline{2-3}
            & T + A (JE) &0.6919 \\ \cline{2-3}
            & T + V (JE) &0.7094\\ \cline{2-3}
            & A + V (JE) & 0.6580\\ \cline{2-3}
            & No Face & 0.7175\\ \cline{2-3}
            & No Body & 0.7286\\ \cline{2-3}
            & T + A + V & 0.7491\\ \cline{2-3}\hline
    \end{tabular}}
    \vspace{-7mm}
    \end{table}

\begin{table}[t]
\centering
\caption{Ablation Study using w-F1 Score}
\label{tab:ablation}
\centering
\smaller {
\begin{tabular}{|l|l|l|l|}
\hline
\textbf{Model}           & \textbf{Modalities} & \textbf{MELD} & \textbf{IEMOCAP} \\ \hline
No Multimodal Attention  & T+A+V               & 0.6160            & 0.7527               \\ \hline
Equal Fusion Weights     & T+A+V               & 0.6934            & 0.7165               \\ \hline
\multirow{3}{*}{No Guided NCE} & No Body             & 0.6343            & 0.6491               \\ \cline{2-4} 
                         & No Face             & 0.6021            & 0.6182               \\ \cline{2-4} 
                         & T+A+V               & 0.7041            & 0.7286               \\ \hline
\textit{AM$^2$-EmoJE}                    & T+A+V               & 0.7198            & 0.7491               \\ \hline
\end{tabular}
\vspace{-6mm}
}
\end{table}

\vspace{5mm}

\section{Conclusion}

\vspace{-2mm}
We present \textit{AM$^2$-EMOJE} with our weighted multimodal attention and query adaptive fusion, allowing us to effectively combine the information from various modes to make better decision about the emotional state of subjects in a group conversation. The model is also trained with our proposed \textit{Guided NCE} loss that allows the model to learn representation of the subjects with only their facial features or body language, allowing us to better preserve the privacy of participants and still achieve performance that is very close to the state-of-the-art. Furthermore, we also propose an effective \textit{Multimodal Joint-Embedding} scheme that allows the model to effectively compensate for missing modalities during inference, allowing its performance on a subset of modalities to be close to its performance on the full set of modalities as shown in our comparison with other state-of-the-art models and in the ablation studies in the absence of Joint-Embedding.

\vspace{-6mm}
\bibliographystyle{IEEEbib}
\bibliography{icme2023template}

\end{document}